# Multimodal data matters: language model pre-training over structured and unstructured electronic health records

Sicen Liu, Xiaolong Wang, Yongshuai Hou, Ge Li, *Member IEEE*, Hui Wang, Hui Xu, Yang Xiang, *Member IEEE*, Buzhou Tang, *Member IEEE*

*Abstract*—As two important textual modalities in electronic health records (EHR), both structured data (clinical codes) and unstructured data (clinical narratives) have recently been increasingly applied to the healthcare domain. Most existing EHR-oriented studies, however, either focus on a particular modality or integrate data from different modalities in a straightforward manner, which usually treats structured and unstructured data as two independent sources of information about patient admission and ignore the intrinsic interactions between them. In fact, the two modalities are documented during the same encounter where structured data inform the documentation of unstructured data and vice versa. In this paper, we proposed a Medical Multimodal Pre-trained Language Model, named MedM-PLM, to learn enhanced EHR representations over structured and unstructured data and explore the interaction of two modalities. In MedM-PLM, two Transformer-based neural network components are firstly adopted to learn representative characteristics from each modality. A cross-modal module is then introduced to model their interactions. We pre-trained MedM-PLM on the MIMIC-III dataset and verified the effectiveness of the model on three downstream clinical tasks, i.e., medication recommendation, 30-day readmission prediction and ICD coding. Extensive experiments demonstrate the power of MedM-PLM compared with state-of-the-art methods. Further analyses and visualizations show the robustness of our model, which could potentially provide more comprehensive interpretations for clinical decision-making.

This work was supported by National Key R&D Program of China (2021ZD0113402 ), National Natural Science Foundations of China (61876052, U1813215 and 62106115), National Natural Science Foundation of Guangdong, China (2019A1515011158), Major Key Project of PCL (PCL2021A06), Strategic Emerging Industry Development Special Fund of Shenzhen (20200821174109001), and Pilot Project in 5G + Health Application of Ministry of Industry and Information Technology & National Health Commission (5G + Luohu Hospital Group: an Attempt to New Health Management Styles of Residents). *(Corresponding author: Yang Xiang, Buzhou Tang)*

Sicen Liu, Buzhou Tang are now with the Department of Computer Science, Harbin Institute of Technology (Shenzhen), Shenzhen, China, and Peng Cheng Laboratory, Shenzhen, China(email: liusicen@stu.hit.edu.cn; tangbuzhou@gmail.com)

Xiaolong Wang is now with the Department of Computer Science, Harbin Institute of Technology (Shenzhen), Shenzhen, China (email: wangxl@insun.hit.edu.cn).

Yongshuai Hou and Yang Xiang are now with Peng Cheng Laboratory, Shenzhen, China (email: houysh@pcl.ac.cn; xiangy@pcl.ac.cn).

Ge Li is now with Peking University and Peng Cheng Laboratory, Shenzhen, China (email: geli@pku.edu.cn).

Hui Wang and Hui Xu are now with Gennlife(Beijing) Technology Co Ltd, Beijing, China (email: wanghui@gennlife.com;xuhui@gennlife.com).

## I. INTRODUCTION

THE growing availability of large-scale electronic health records (EHR) plays an essential and determinant role in data-driven clinical decision support systems [1]–[3], providing more opportunities to improve healthcare by using artificial intelligence methods [4]–[9]. There have been an extensive array of successes achieved using a nascent technique, named deep learning, in optimizing healthcare [10]–[15], triggering more in-depth analyses on the nature of EHRs to develop effective models [16]–[21]. With the advances of pre-trained language models (PLM) for deep learning, more impressive performance on downstream tasks have been obtained [22]–[27].

The data from EHR can be generally categorized into well-organized structured data (e.g., clinical codes) and free-style unstructured data (e.g., clinical narratives). They carry information on patients' health status and different stages of medical care [28]–[30], providing the basis for physicians' decision-making [31], [32]. On the one hand, although relevant codes are usually manually assigned by clinical practitioners, e.g., for billing use, they may suffer from the incompleteness problem when considering more comprehensive purposes. For example, for diabetes mellitus (DM) patients, the type 2 diabetes mellitus (T2DM) could be contaminated with type 1 diabetes mellitus (T1DM) subjects because many patients are assigned the code for *diabetes mellitus, unspecified type* [33]. Further, although the administrative coding systems translate healthcare diagnoses and medication records into universal codes, they cannot provide a granular view of a patient's presentation, disease severity and clinical sequence during an episode of medical care [34]. Free text may be chosen when no code precisely describes clinical findings or when there is a need to give supporting evidence for a diagnosis or suspicion [35]. On the other hand, multi-source structured data contain more health conditions of patients, including medication information from the clinical prescription records, that may not be included in the clinical notes. Hence, physician usually need to combine clinical notes and codes to improve the phenotyping ability of EHRs. In this paper, we name these structured and unstructured data as multimodal data, following the definitions in previous work [36].

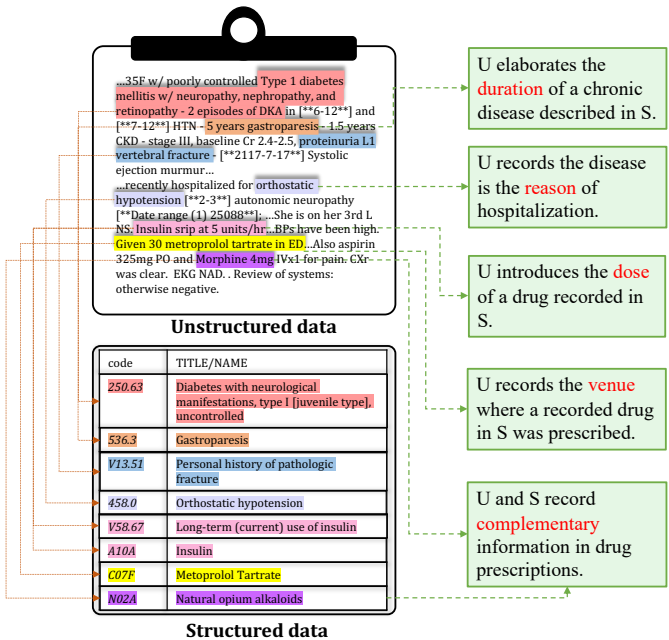

Fig.1. An example of paired multimodal data in a visit record of a patient. Structured data included codes that were collected and summarized according to practitioners based on the patient's clinical status. The unstructured data recorded more narrative information, such as the duration of a particular disease, the primary reason for hospitalization, and the dosage of a prescription from clinicians. The green boxes show the complementation of the unstructured and structured data. U: unstructured data, S: structured data.

The data from each modality not only convey informative messages of patients alone but also can be comprehensively overlapped and correlated [37]–[39]. Figure 1 shows an example of the interactive relationships in a paired piece of multimodal EHR. We can observe that there exist different types of interactions between the two modalities (shown in the green rectangles in Figure 1), which demonstrate the complementarity between data from different modalities.

However, most existing studies in informatics only focus on one modality [40]–[52] or use straightforward strategies (e.g., concatenation) to combine different modalities [53]–[58]. In general, most of them have limitations in modeling the interactions in multimodal data, either ignoring their complementary relationship or insufficient in modeling the complex interaction between multimodal EHR data. Taking the record in Figure 1 as an example, if a model only learns from the structured data, the fact that *the disease gastroparesis has been lasted for 5 years* would never be captured, the details of which are recorded in the unstructured part. Furthermore, the medication code *N02A (natural opium alkaloids)* in the unstructured part can include several types of alkaloids, while the narratives clarify this is *morphine 4mg*.

In this paper, we hypothesize that the inherent connection between the two modalities are critical and could be captured by modeling the multimodal interactions. We proposed a Medical Multimodal Pre-trained Language Model (MedM-PLM) to build the connections between multimodal EHRs. In detail, two Transformer-based neural network components are firstly adopted to learn representative characteristics from each modality, which could inherit the advantages of unimodal PLMs that were well pre-trained to sustain their modal-specific properties. The unimodal module is simple and flexible that any structured code- and unstructured text-based PLMs can be plugged in. A cross-modal module is further developed to model the interaction relationships between modalities. In the cross-modal module, we designed two pre-training tasks, i.e., Text-to-Code and Code-to-Code, to pre-train MedM-PLM, so that downstream clinical tasks could benefit from the learned representations.

We evaluated the MedM-PLM model on three medical prediction tasks: medication recommendation, 30-day readmission prediction, and ICD (International Classification of Diseases) coding, which are popular multimodal tasks over EHRs. Different benchmark and state-of-the-art methods were compared through extensive experiments, and the results demonstrate the effectiveness of the proposed MedM-PLM. Experiments on few-shot learning scenarios further show the scalability of the pre-trained model, indicating the potential of MedM-PLM in applying to new or rare diseases.

Our primary contributions are summarized as follows:
(1). We proposed a novel multimodal PLM for jointly modeling unstructured data and structured data in EHRs, which can learn cross-modal interactions while retaining unimodal representation capacities;
(2). We conducted fine-tuning experiments on three medical prediction tasks to evaluate the performance of our proposed model. The results demonstrate the power of MedM-PLM in utilizing multimodal EHRs;
(3). Experiments with different training proportions show that our model performs consistently better than the baselines, even on small training sets;
(4). We have made our codes and pre-trained model publicly available to enhance reproducibility, which could be beneficial for a broader range of researchers.

## II. METHODS

### A. Data Preparation

We used a large and publicly available database, the Medical Information Mart for Intensive Care III (MIMIC-III) dataset[1], in our experiments. The dataset contains EHR data associated with 53,423 distinct hospital admissions by 35,164 adult patients (age 16 years or above) between 2001 and 2012. The data include vital signs, medications, procedure codes, diagnostic codes, and clinical narratives from physicians and practitioners during patients' hospitalization. We selected patients associated with both unstructured and structured data to ensure the completeness of multimodal data and for fair comparison. The description of the MIMIC-III dataset is summarized in Table I.
In the pre-training phase, we used 80% of the single visit records (extracted from both single-visit patients and multi-visit patients in selected patients) as the training set. In detail, we used the discharge summary record corresponding to the patient's structured data as the unstructured counterpart. We followed ClinicalBERT [22] to preprocess the unstructured data,

---
[1] https://mimic.mit.edu/

in which words were converted to lowercase, and line breaks and carriage returns were removed.

TABLE I
STATISTICS OF THE MIMIC-III DATASET

| Characteristic | Number |
|---|---|
| Total patients | 35,164 |
| Single-visit patients | 29,734 |
| Multi-visit patients | 5,159 |
| Selected patients | 33,413 |
| Total diagnoses | 6,646 |
| Avg # of diagnoses | 11.11 |
| Total medications | 155 |
| Avg # of medication | 9.23 |

We followed the configuration of G-BERT [25] for the medication recommendation task in the fine-tuning phase, in which all the multi-visit sequences were selected to form the total dataset, and divided the dataset of the selected multi-visit patients with the ratio of 8:1:1. For the 30-day readmission prediction task, we made statistics on the readmission label of each visit and set those samples where the patient's next visit time is within 30 days as positive whereas the others as negative samples. The positive and negative samples were sub-sampled with a 1:1 ratio from the total visit records, resulting in 4,660 records for training, 532 for validation, and 534 for testing. For the ICD coding task, we used the dataset analogous to the prior work CAML [59] with simple adaptations. Specifically, we selected the unstructured data following CAML [59] and searched the corresponding structured data in the MIMIC-III dataset to obtain the structured and unstructured data pairs. To align with the input format for pre-training, we set the max token sequence length as 512. The details of the datasets for pre-training and fine-tuning are summarized in Table II.

TABLE II
NUMBER OF SAMPLES FOR PRE-TRAINING AND FINE-TUNING

| Task | Training set | Validation set | Testing set |
|---|---|---|---|
| Pre-training | 39,550 | - | - |
| Medication recommendation | 4,344 | 543 | 543 |
| 30-day readmission | 4,660 | 532 | 534 |
| ICD coding | 8,066 | 1,573 | 1,729 |

### B. Model Overview

MedM-PLM handles the inputs from structured and unstructured EHRs and enhances the EHR representing ability by modeling the interactions between two modalities. Figure 2 shows the architecture of MedM-PLM. Based on the input unstructured text sequence and structured code sequence, we firstly designed a unimodal module to learn data representations from each modality to preserve the modal-specific characteristics. More specifically, a BERT-like component and a G-BERT-like component are leveraged to encode the unstructured and structured inputs respectively. After obtaining the unimodal representation, a cross-modal module is then introduced to integrate the multimodal representations that learn the interaction information between structured and unstructured data. Furthermore, two specific pre-training tasks are designed in the cross-modal module to capture complementary information and model interactions between different modalities, i.e., the Text-to-Code prediction and Code-to-Code prediction tasks.

### C. Notions and Definitions

In EHRs, a patient's record can be represented as a paired unstructured sequence and structured sequence, denoted as $X = [(W_t, C_t)]$, where $t \in (1,2,...,\mathcal{T})$, and $\mathcal{T}$ is the number of visits. $W_t = (w_t^1, w_t^2, \cdots, w_t^k)$ is the word sequence of an unstructured record within a visit, and $C_t = (c_t^1, c_t^2, \cdots, c_t^l)$ is the corresponding structured part. Here, we use $c \in C$ to represent each code, which can be either a diagnosis or a medication code. We used a special symbol [CLS] as the first token of each sequence. The final hidden state corresponds to the special symbol [CLS] is summarized from the sequence representation, which is used as the aggregated information of visit-level [22], [26], [60]. We remove the subscript $t$ from $W_t$ and $C_t$ for simplicity.

### D. Unimodal Module

The unimodal module was designed to make full use of the modality-specific semantics, including two unimodal components. In each component, the corresponding input can be flexibly encoded by any deep learning layers. In this study, we utilized two Transformer-based components for encoding, each of which contains an embedding block followed by multiple Transformer blocks. Specifically, for the structured data component, in the embedding layer, inspired by G-BERT [25], we used two hierarchical ontologies to categorize the medication and diagnosis codes. i.e., Anatomical Therapeutic Chemical, Third Level (ATC-3)[2], and the International Classification of Diseases, Ninth Version (ICD-9)[3]. Both are hierarchical with tree structures, and each medical code is firstly represented as a leaf node in each ontology tree. Graph Attention Networks (GAT) [61] is then applied to aggregate the representations of each node and its direct children nodes to obtain the enhanced node representation $e_a$ for each non-leaf node. Further, the embedding of each leaf node $e_c$ is updated by fusing the message passed from the ancestor nodes, implying a broader range of medical code information.

$$e_{NL} = GAT(A_{NL}, M) \quad (1)$$

$$e_c = GAT(A_L, e_{NL}) \quad (2)$$

where $A_{NL}$ is the adjacent matrix (directed graph) of the non-leaf nodes, $M$ is the initial embedding matrix of nodes, and $A_L$ is the adjacent matrix of the leaf nodes.

For the unstructured data component, in the embedding block,

---
[2] https://www.whocc.no/atc/structure_and_principles/
[3] https://www.cdc.gov/nchs/icd/icd9.htm

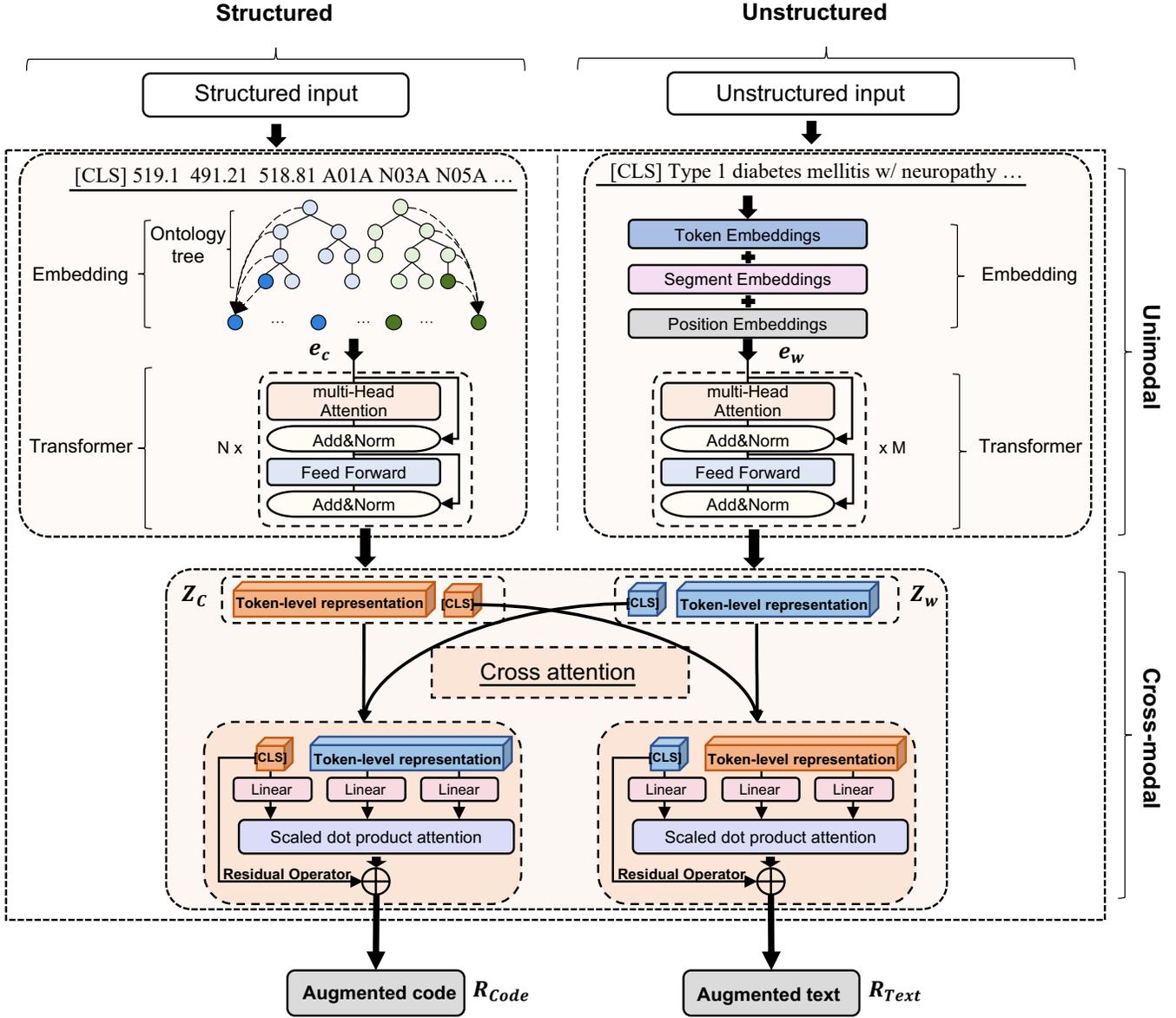

Fig. 2. The model architecture of MedM-PLM. It includes a unimodal module that contains the unstructured text component and the structured code component (the upper part), and a cross-modal module (the bottom part).

the representation of each visit sequence is generated by summing up the token embeddings, segment embeddings, and position embeddings for each token, which is similar to BERT [60].

$$e_w = SUM(e_{w_{token}}, e_{w_{segment}}, e_{w_{position}}) \quad (3)$$

Through the embedding block of each other, we obtained the unstructured text sequence representation $e_w$ and structured code sequence representation $e_c$ for each visit. The multilayer Transformer [62] architecture was further employed as the visit encoder for each modality.

$$Z_W = Encoder(e_w, \theta_w) \quad (4)$$

$$Z_C = Encoder(e_c, \theta_c) \quad (5)$$

where $\theta_w$ and $\theta_c$ are learnable parameters. Through the unimodal module, we obtained both the visit-level and token-level representations of each modality.

### E. Cross-modal Module

After obtaining the modal-specific representation, we designed a cross-attention mechanism to integrate the information from multi-modalities and a residual operator to augment model-specific and cross-modal information. The cross-attention mechanism is used to learn the intrinsic interaction between structured and unstructured data. The attention mechanism automatically captures the relative semantic information of each modality and assigns higher weights to correlated segments. Specifically, we used the visit-level representation from the structured data as a query to collect the relevant information from the unstructured data. Meanwhile, we performed the same attention-based query from the unstructured data to collect the structured data information from the structured data. The cross-attention and residual operations are defined as:

$$R_{Code} = Softmax\left(\frac{\tilde{Z}_C Z_W^T}{\sqrt{d_{Z_W}}}\right) Z_W + \tilde{Z}_C \quad (6)$$

$$R_{Text} = Softmax\left(\frac{\tilde{Z}_W Z_C^T}{\sqrt{d_{Z_C}}}\right) Z_C + \tilde{Z}_W \quad (7)$$

where $\tilde{Z}_C$ and $\tilde{Z}_W$ are the visit-level embeddings, $Z_C$ and $Z_W$ are the embeddings of each code and word. Our cross-modal module finally generates the augmented visit-level code representation $R_{Code}$ and text representation $R_{Text}$.

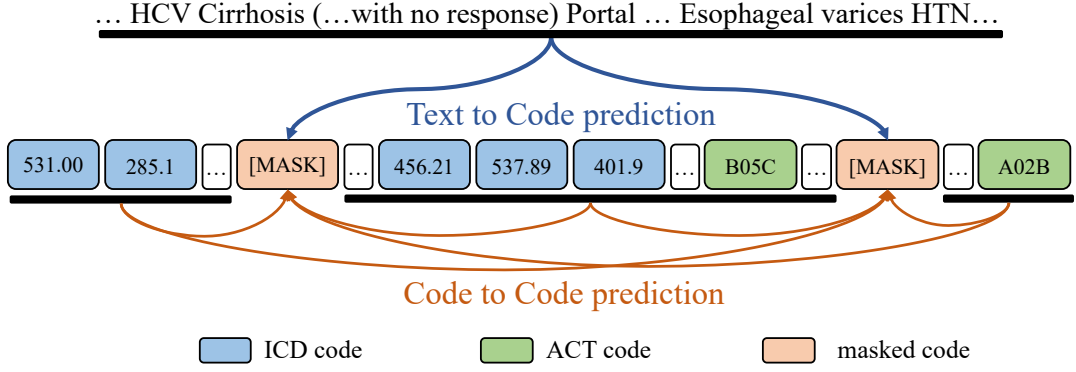

Fig.3. Graphical illustration of the pre-training tasks.

### F. Pre-training

Inspired by ALBEF [63], we modified the pre-training task Masked Language Model (MLM) into a multimodal version. In particular, we defined two pre-training tasks, i.e., Text-to-Code and Code-to-Code (Figure 3), meaning to re-construct each code using the paired free text and re-construct each code using the surrounding codes. These modified pre-training tasks aim to learn the visit-level context-aware semantics of structured data from structured and unstructured sequences. The pre-training objectives are:

$$\mathcal{L}_{T2C} = -\log(C_{[mask]}|R_{Text}) \quad (8)$$

$$\mathcal{L}_{C2C} = -\log(C_{[mask]}|R_{Code}) \quad (9)$$

where $C_{[mask]}$ is the masked token from the inputs. $L_{T2C}$ and $L_{C2C}$ are cross-entropy losses.

The final objective function is the sum of the above two losses:

$$\mathcal{L} = \mathcal{L}_{T2C} + \mathcal{L}_{C2C} \quad (10)$$

In the unimodal module, for the pre-training of unstructured data, we initialized the model with pre-trained parameters from ClinicalBERT [22]. For the structured data, the parameters were randomly initialized due to different vocabularies from G-BERT [25]. For the masking strategy of the two pre-training tasks, we followed the masking strategy of BERT [60], which randomly selected 15% tokens from the structured code sequence to mask and used an 80% rate to replace the selected code by [MASK], a 10% rate to keep the code not changing, and a 10% rate to change to a random code.

### G. Evaluation

After obtaining the pre-trained multimodal representations, MedM-PLM can be applied to downstream tasks through fine-tuning to improve the performance of these tasks. By adding task-adaptive classification layers, the downstream tasks could be binary classification (e.g., 30-day readmission prediction), multi-label classification (e.g., medication recommendation), and else. We conducted experiments on three fine-tuning clinical prediction tasks, and the datasets were also extracted from MIMIC-III.

#### 1) Finetuning Task 1: Medication recommendation

Medication recommendation is an important application in healthcare, which aims to automatically recommend medications (drugs) that are suited for a patient's health condition [16], [64]. This task is defined as a multi-label prediction task that utilizes the historical data records to predict the drug sequence of the next visit. Similar to G-BERT, we utilized the diagnosis codes and medication codes from the historical visit records. Further, we concatenated the mean of augmented visit-level representations of diagnoses and medications in the historical records (0 to $t-1$th visits) and the augmented representation of diagnoses for the $t$-th visit. For the text, we concatenated the mean of augmented visit-level text representations of historical records (0 to $t-1$th visits) to predict the drug of the $t$-th visit.

We built a multi-label Multi-Layer Perceptron (MLP) as the predication layer:

$$y'^d_t = E[R_{Text_i}|i<t] \parallel E[R_{Code_i}|i<t] \parallel E[R_{C_d}|j=t] \quad (11)$$

$$y^d_t = Sigmoid\left(MLP(y'^d_t)\right) \quad (12)$$

where $E$ is the expectation function, "$\parallel$" is the concatenation operator, and $C_d$ is the diagnosis sequence. Given the ground truth labels $\hat{y}^d_t$ of each timestamp, the loss function is:

$$\mathcal{L}_d = -\frac{1}{T-1}\sum_{t=2}^{T}(\hat{y}^d_t \log(y^d_t) + (1-\hat{y}^d_t)\log(1-y^d_t)) \quad (13)$$

## 2) Fine-tuning Task 2: 30-day readmission prediction

The prediction of 30-day readmission is meaningful in practice in improving patients' life quality and lowering down the financial cost. The task considers a patient encounters readmission if the admission date of the patient was within 30 days after the discharge date of the previous hospitalization, and thus is a binary classification task. For each visit, we concatenated the pre-trained enhanced unstructured text feature $R_{Text}$ and structured code feature $R_{Code}$, and then used a MLP to generate the final output of the current visit:

$$y_r = Sigmoid(MLP(R_{Text} \parallel R_{Code})) \quad (14)$$

$$\mathcal{L}_r = -\sum_{p=1}^{P} \hat{y}_r log(y_r) + (1 - \hat{y}_r)\log(1 - y_r) \quad (15)$$

where $\hat{y}_r$ is the label of readmission.

## 3) Finetuning Task 3: ICD coding

ICD coding for large-scale clinical notes is labor-intensive and error-prone, while machine learning methods could help automatically reduce time and laborious cost [65]. ICD coding usually is treated as a multi-label classification problem, in which relevant ICD codes to the patient records are assigned automatically. For the multimodal EHR input, we also used the medication information in the corresponding visit as a complement to ICD coding. After the pre-training phase, we concatenated the outputs from the cross-modal module, which represents the visit-level unstructured text and structured code representations of the current record. Then an MLP classification layer is added to generate the ICD codes:

$$y_I = Sigmoid(MLP(R_{Text} || R_{C_m})) \quad (16)$$

where $C_m$ is the medication sequence.

The training objective is to minimize the binary cross-entropy loss between the prediction $y_I$ and the target $\hat{y}_I$:

$$\mathcal{L}_I = -\sum_{j=1}^{J} \hat{y}_I log(y_I) + (1 - \hat{y}_I)\log(1 - y_I) \quad (17)$$

## III. EXPERIMENTS

### A. Baselines

We compared MedM-PLM with the following baselines. All deep learning methods are implemented in PyTorch [66].

**LR:** Logistic Regression is a conventional machine learning method [67]. Compared in the medication recommendation task by building binary one-versus-rest classifiers, LR uses the binary relevance method to perform multi-label classification. This method involves training one binary classifier independently for each label[4]. LR was also compared in the 30-day readmission task by building a binary classifier.

**RNN:** Recurrent Neural Network [68] uses a patient record sequence as input to learn the hidden representation of the patient and performs the binary classification based on the hidden states.

**CNN:** The one-dimensional Convolutional Neural Network [69] was employed to learn text representation for the 30-day readmission prediction and ICD coding tasks, while the 30-day readmission prediction was treated as a binary classification and the ICD coding task was treated as a multi-label classification task.

**G-BERT:** G-BERT combines GNNs and BERT for medical code representation, in which GNNs are used to represent the hierarchical structures of medical codes. The GNN representations are further integrated into a Transformer-based pre-trained model [25].

**Med-BERT:** Med-BERT adapts the BERT framework for the natural language processing domain to structured EHRs, which defines serialization embeddings to denote the relative order of each code [27].

**ClinicalBERT:** ClinicalBERT was pre-trained BERT [60] using clinical notes and fine-tuned for the task of hospital readmission prediction [22].

**G-BERT+ClinicalBERT:** We directly concatenated the structured code representation of G-BERT and the unstructured data representation of ClinicalBERT. This method was used to verify the different combination ways for multimodal data.

**Med-BERT+ClinicalBERT:** Similar to G-BERT +ClinicalBERT, Med-BERT+ClinicalBERT concatenated the pre-trained representations of Med-BERT and ClinicalBERT.

### B. Implementation Details

In the unimodal module, for the unstructured data component, we used 12 transformer blocks, 12 attention heads, and a hidden dimension of $768 (L = 12, H = 768, A = 12)$. For the structured data component, we used 2 encoder layers, 2 attention heads, and a hidden size of 300. The ontology embedding size is 75, and the number of heads for ontology aggregation attention is 4. In detail, we set the maximum sequence length of unstructured data as 512. Since we did not use the Text-to-Text MLM, we froze the previous ten layers of the encoder and the embedding layer of ClinicalBERT and only optimized the parameters of the last two encoder layers to better inherit the pre-trained parameters from ClinicalBERT and align with the structured data component. The maximum sequence length for the structured data was set as 61, which is the maximum number of codes in a single visit. We masked the structured data using a 15% rate similar to the original BERT [60]. We used the learning rate of $5e − 4$ and dropout rate of 0.1, and the training batch size of 32. Pre-training was done using the Adam [70] optimizer. The model was pre-trained on the corresponding dataset with a maximum of 200 epochs. Two GeForce RTX 3090 GPUs were leveraged to pre-train the MedM-PLM model, and the early-stopping method was utilized. In the fine-tuning phase, we set different learning rates for the tasks, i.e., $5e − 5$ for medication recommendation, $2e − 5$ for 30-day readmission prediction, $1e − 5$ for ICD coding, and $3e − 5$ for NER (details in V. DISCUSSION). All evaluations were duplicated five times with different random seeds to reduce overfitting, and the average values and standard deviations of the evaluation metrics were reported.

---
[4] https://scikitlearn.org

TABLE III
SUMMARY OF PERFORMANCE ON THREE DOWNSTREAM TASKS ON F1, ACCURACY, AND AUC. THE STANDARD DEVIATIONS ARE LISTED IN BRACKETS

| Task | Model | F1% | Accuracy% | AUC% |
|---|---|---|---|---|
| Medication recommendation | LR(Code) | 61.49(0) | 89.07(0) | 77.43(0) |
| | RNN [68] (Code) | 58.48(0.03) | 90.55(0.01) | 91.98(0.02) |
| | G-BERT [25] (Code) | 65.38(0.09) | 91.69(0.04) | 94.38(0.02) |
| | Med-BERT [27] (Code) | 61.47(0.41) | 91.02(0.01) | 93.04(0.14) |
| | Med-BERT+ClinicalBERT | 61.41(0.03) | 90.93(0.05) | 92.97(0.04) |
| | G-BERT+ClinicalBERT | 65.37(1.48) | 91.53(0.31) | 94.39(0.33) |
| | MedM-PLM$_{-cross\_modal}$ | 66.17(0.10) | 92.03(0.01) | 94.42(0.04) |
| | MedM-PLM | **70.21(0.08)** | **92.93(0.03)** | **95.57(0.01)** |
| 30-day readmission prediction | CNN [69] (Text) | 57.84(1.70) | 62.55(1.37) | 66.74(1.02) |
| | ClinicalBERT [22] (Text) | 63.86(1.41) | 64.19(1.40) | 69.37(1.43) |
| | LR(Code) | 63.17(0) | 65.73(0) | 65.73(0) |
| | G-BERT [25] (Code) | 64.82(1.02) | 65.42(0.64) | 69.57(0.43) |
| | Med-BERT [27] (Code) | 64.52(0.78) | 64.22(0.97) | 69.06(1.61) |
| | ClinicalBERT+G-BERT | 65.63(1.12) | 65.67(1.11) | 70.79(0.33) |
| | ClinicalBERT+Med-BERT | 64.35(0.85) | 64.48(0.92) | 69.38(1.26) |
| | MedM-PLM$_{-cross\_modal}$ | 61.54(1.57) | 66.54(0.87) | 71.23(0.83) |
| | MedM-PLM | **68.61(0.83)** | **68.77(069)** | **74.70(0.50)** |
| ICD coding | CNN [69] (Text) | 49.08(0.73) | 32.53(0.64) | 84.06(0.68) |
| | ClinicalBERT [22] (Text) | 49.72(1.80) | 33.10(1.59) | 84.11(0.44) |
| | ClinicalBERT+G-BERT(Drug) | 50.14(0.55) | 33.46(0.49) | 85.86(0.21) |
| | MedM-PLM$_{-cross\_modal}$ | 51.51(0.47) | 34.07(0.03) | 86.70(0.04) |
| | MedM-PLM | **52.09(0.65)** | **35.22(0.60)** | **87.46(0.05)** |

*MedM-PLM$_{-cross\_modal}$ means removing the multimodal module from our model. The parentheses (Code) means using the structured data as input, and (Text) means using the unstructured data as input, and (Drug) means only using the drug code as the input of the structured component.

## IV. RESULT

Table III presents the results on the three downstream tasks with the best value of each column boldfaced. The primary evaluation metric of the three tasks is Area Under the Receiver Operating Characteristic (AUC). We also listed the accuracies and F1s. From Table III, we can generally draw the following conclusions: 1) The deep learning-based methods perform much better than conventional machine learning-based methods; 2) Methods with the addition of PLMs have better performances than those without; 3) Combining structured and unstructured data do not always obtain better results than using only a uni-modal input; 4) Using the cross-modal module in MedM-PLM is superior to that does not use; 5) Using MedM-PLM obtains the best result in all tasks.

For example, in the medication recommendation task, G-BERT and Med-BERT outperform LR and RNN when only code is taken as the input. When combining structured code and unstructured text, however, the direct concatenation methods, i.e., Med-BERT+ClinicalBERT and G-BERT+ClinicalBERT, perform worse than G-BERT. In comparison, our model without modeling multimodal interactions (MedM-PLM-cross_modal) performs comparably with G-BERT. And when adding the cross-modal module, MedM-PLM improves G-BERT by 1.15%. In readmission prediction, using structured code only or unstructured text only can both achieve an AUC over 0.69, and ClinicalBERT+G-BERT with multimodal input slightly outperforms them. MedM-PLM-based models further improve the AUCs, and MedM-PLM even improves ClinicalBERT+G-BERT by 3.91%. Similar trends can also be observed in the task of ICD coding. Besides, in the three tasks, the performance of MedM-PLM-cross_modal is far worse than the MedM-PLM, which also illustrates the effectiveness of the cross-modal module.

Further, in order to verify if MedM-PLM can be beneficial in different cases, especially in scenarios with smaller training data, we fine-tuned the model over various training proportions by setting different training ratios. In Figure 4, the broken line charts show that MedM-PLM consistently outperforms other baselines. Even in the extreme circumstance where only 10% of the training set is available to train the downstream model, MedM-PLM still shows its superiority in contributing to all prediction tasks. In the ICD coding task in Figure 4(c), the pre-training model ClinicalBERT improves CNN by almost 5% when training on 10% of the training data, and MedM-PLM further improves ClinicalBERT by about 3%. We can also observe from Figure 4(a) and 4(c) that models without PLMs (e.g., RNN, CNN) have poor performances when the training size is extremely small (e.g., 10%).

## V. DISCUSSION

Upon analyzing the results in Table III, we can conclude that the PLM-based methods obtain much better results than conventional machine learning and deep learning methods, e.g., LR and RNN, in general, and the results also verify that the modeling of multimodal data using our proposed MedM-PLM is effective across different tasks. Further experiments also demonstrate its stability on different sizes of the training set in the fine-tuning phase. The success of MedM-PLM can be attributed to that the pre-training phase well captures the complex and interactive semantics of multimodal EHRs through two unsupervised pre-training tasks, i.e., Text-to-Code and Code-to-Code prediction. Using Text-to-Code, the context of each code can be expanded by more detailed descriptions of the code or other correlated codes. For example, in Figure 1, the clinical narratives in the purple box *Morphine 4mg* can be modeled as a piece of

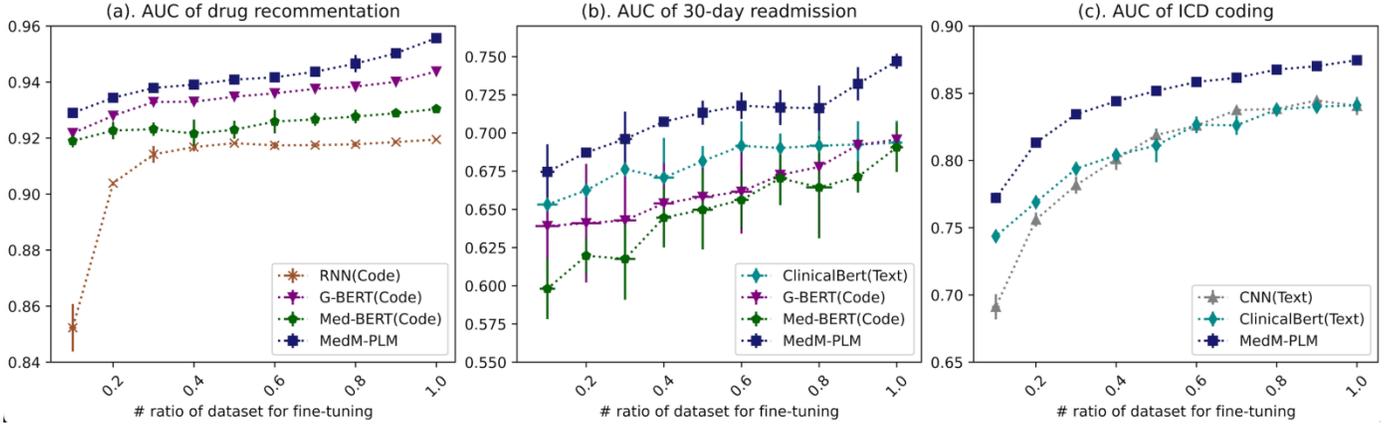
Fig.4. Comparisons of different models by using different training ratios.

evidence for the coding of *N02C* (*Migraine medication*). Using Code-to-Code[5], the dependencies between different medical codes can be further enforced, e.g., drugs and diagnoses. We did not use Text-to-Text prediction since this process has already been well exploited in the pre-trained ClinicalBERT, and we did not observe any improvements during our preliminary attempts. The unsupervised pre-training tasks aim to re-construct the masked tokens using information from different modalities, thus can help the model collect intrinsic relationships among multimodal data, which is more powerful than direct concatenation, e.g., G-BERT+Clinical-BERT.

As mentioned above, using multimodal data does not always outperform unimodal methods. For example, in the medication recommendation task, the AUCs of G-BERT+ClinicalBERT, Med-BERT+ClinicalBERT, and MedM-PLM-cross_modal are lower or only comparable with the unimodal pre-trained model G-BERT. We believe this is partly determined by the characteristics of different scenarios. For example, in medication recommendation, the basic predictions are inferred from the historical structured data, in which the diagnosis codes and historical medication codes reflect the health situation. However, the unstructured data contain not only related symptoms but also some extra information, e.g., the family history and social history of a patient, which might add noise to the prediction if an effective information refinement is not involved.

We can infer from Table III that there are several possible reasons why MedM-PLM can outperform the other methods: 1) By comparing conventional machine learning-based methods with our proposed MedM-PLM, we deem that the prior knowledge learned from the pre-training phase could have been effectively inherited. 2) By comparing unimodal PLMs with our proposed MedM-PLM, the multimodal information has been added to the patient representation to enhance the model's representation capacity. Further, MedM-PLM uses well-designed pre-training tasks to model the multimodal interactions, which fills the gap between multimodal PLMs. 3) By comparing G-BERT and MedM-PLM in the medication recommendation task, we notice that the length of the predicted drug sequence generated by MedM-PLM is closer to the ground truth, and the recall is higher than that using the unimodal input. This phenomenon shows adding the expanded information from unstructured data may have added constraints in helping generate the drug sequence. 4) By comparing the straightforward concatenation of two PLM representations with our proposed MedM-PLM, we find that the cross-modal module is helpful in capturing the inherence correlation automatically. Thus, our proposed MedM-PLM is able to learn more effective and robuster representations.

To more intuitively understand that MedM-PLM is concerned with the most informative multimodal information, we selected a case and visualized the model's attention weights using a heat map, which shows the focus of the model through highlighting the most informative words. According to Figure 5, MedM-PLM can automatically assign variant weights to words in the unstructured data and codes in the structured data that have different importances in determination. These words and codes might be either corresponding, e.g., *edema* with *348.5* and *parasagittal meningioma* with *225.2*, or complementary, e.g., *Ativan IV* with *N05A* (*antipsychotics*) and *Fosphenytoin* with *N03A* (*antiepileptics*). It is more like mimicking the clinicians who can recognize the critical factors underlying according to the EHR of a patient during the decision-making process. Following this information, the EHR of a patient can be automatically tagged with important cross-modal signals identified, and better visualizations and interpretations can be provided.

In summary, MedM-PLM can effectively model the interaction between structured data and unstructured data while preserving the modal-specific representation capacity of the unimodal data. The most informative segments were concentrated in MedM-PLM through the cross-modal module. Furthermore, the pre-trained model has shown its robustness and potential value in clinical decision-making, where solid performances on a variety of downstream tasks have been achieved.

Furthermore, Biomedical Named Entity Recognition (NER)

---
[5] We added Code-to-Code prediction since the vocabulary of G-BERT is different from ours and we need to pre-train the parameters of the G-BERT component from scratch.

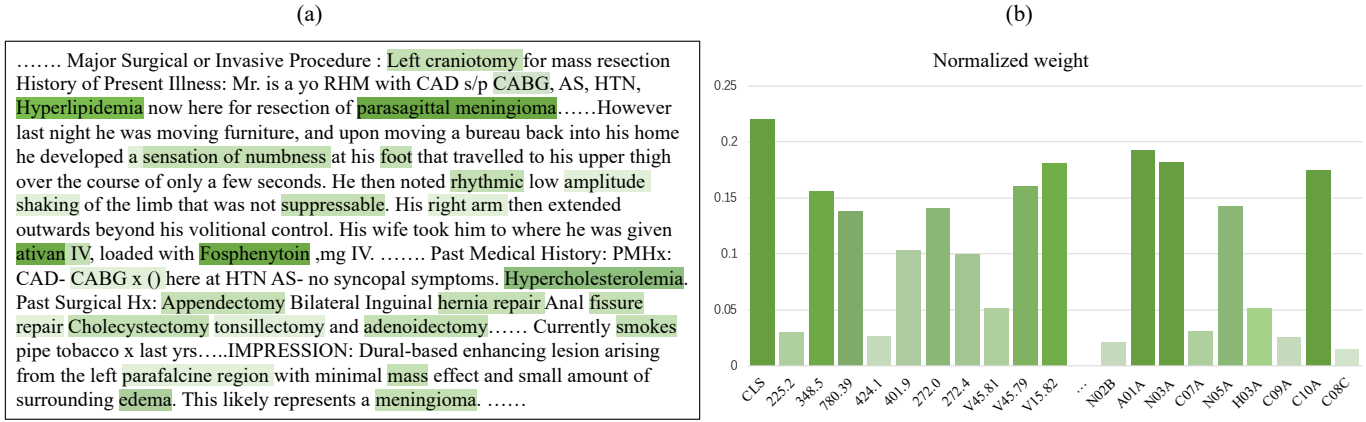

Fig. 5. Visualization of attention for the text fragment and the code sequences. The gradient of the color indicates the degree of importance.

was taken as an auxiliary task to validate if the multimodal pre-trained model could also perform well on a unimodal task. We adopted the concept extraction task of the 2010 i2b2/VA Workshop on Natural Language Processing Challenges for Clinical Records [71]. The details of the dataset are shown in Table IV.

TABLE IV
DETAILS OF 2010 I2B2 NER DATASET

| | NER (2010-i2b2) | | |
|---|---|---|---|
| Dataset | Training set | Validating set | Testing set |
| #Sentence | 14,803 | 1,512 | 27,625 |

The NER task focuses on extracting medical concepts from patients' reports, which is purely based on unstructured data. We performed the NER task based on the BiLSTM-CRF [72] framework and replaced the embedding layer with the output of the unstructured data component of MedM-PLM. The primary evaluation metric is strict F1 score.

TABLE V
PERFORMANCE OF NER TASK

| Task | Model | F1% |
|---|---|---|
| | biLSTM_CRF [72] | 83.81 |
| NER | biLSTM_CRF+ClinicalBert | 85.77 |
| | **biLSTM_CRF+MedM-PLM** | **86.29** |

As shown in Table V, comparing the results of biLSTM_CRF+UMM-PLM and biLSTM_CRF+ClinicalBERT, MedM-PLM still maintains a good representation capacity.

TABLE VI
PERFORMANCE OF MEDICATION RECOMMENDATION TASK OF ABLATION EXPERIMENT

| Task | Model | F1% | Accuracy% | AUC% |
|---|---|---|---|---|
| Medication recommendation | G-BERT[25] | 65.75(0.33) | 91.74(0.07) | 94.40(0.06) |
| | **G-BERT(MedM-PLM)** | **67.84(0.12)** | **92.23(0.04)** | **95.15(0.02)** |

We also replaced the model parameters of G-BERT with pre-trained parameters of the MedM-PLM structured data modality component to verify whether the special pre-training task was beneficial to the structured data. The results are shown in Table VI. From Table VI, we can find that the performances of G-BERT are improved by using the pre-trained parameters from MedM-PLM. And these results demonstrate the effectiveness of MedM-PLM.

There are also several limitations of the current study. Firstly, we only selected medications and diagnoses for the structured data part and ignored others such as vital signs, laboratory measures, and procedure codes, which are also informative but might need more expertise to preprocess. Secondly, the MedM-PLM model only uses the single-visit record in the pre-training phase. This will cause the missing of the time-series information and the continuity of EHRs, which is also partly due to the nature of the MIMIC-III data. Thirdly, we truncated the length of each unstructured data sequence into a fixed number, which limited the content of the unstructured text information. Furthermore, the clinical text may include irrelevant information, misspellings and unstandardized abbreviations, which may mislead the learning process. We will probe more data de-noise strategies in our future work. Lastly, the notes of the MIMIC-III dataset are mainly from the intensive care unit, which might not be quite scalable to other clinical records. The efficacy of MedM-PLM that is evaluated on tasks generated from the particular dataset also limits the generalizability of the model. In future studies, we will explore the addition of more clinical variables and temporal patterns to our pre-trained model to improve the scalability and generalizability.

## VI. CONCLUSION

We propose a unified medical multimodal pre-training model named MedM-PLM in this work. The model was pre-trained to capture both unimodal representation abilities and cross-model interactions from EHRs and was evaluated on three downstream tasks. Experiments demonstrate the superiority and stability of the model. We also tested the performance of MedM-PLM on smaller training sets, which further verified the capability of the model in few-shot learning cases. We expect our model could assist in more application scenarios where both structured and unstructured EHRs are available.

## VII. AUTHORS CONTRIBUTIONS

BT, YX initialized the conceptualization of the project. SL, YX, and BT designed the methods. SL led the implementation of the methods, with substantial inputs from YX and BT, YH, HW, and HX. SL conducted the experiments and produced the results. SL and YX led the visualization and led the writing, with substantial inputs from BT and GL. XW, YX, BT and GL supervised the execution of the project.

## VIII. DATA AVAILABILITY

The pre-training dataset MIMIC-III 1.4 is publicly available at https://mimic.physionet.org/. This is a restricted-accessed resource hosted by PhysioNet, which should be used under the license from the credentialed account of PhysioNet. The downstream tasks of medication recommendation, 30-day readmission, and ICD coding are also associated with MIMIC-III. The dataset for the NER task is open-resource and available at https://portal.dbmi.hms.harvard.edu/projects/n2c2-nlp/.

## IX. CODE AVAILABILITY

The codes for the MedM-PLM model and the pre-trained model are shared here: https://git.openi.org.cn/liusc/3-6-liusicen-multi-modal-pretrain.

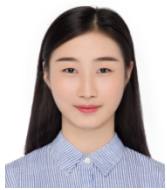

**Sicen Liu** is currently pursuing the Ph.D degree at Harbin Institute of Technology, Shenzhen under the supervision of Prof. Xiaolong Wang in the field of medical informatics.

Her current research interests include sequence learning, medical informatics, and artificial neural networks.

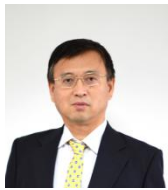

**Xiaolong Wang** received the Ph.D. degree in computer application technology from the Harbin Institute of Technology, Harbin, China, in 1989. He is currently a Professor with Harbin Institute of Technology, China. His research interests include intelligent input method, online finance information platform, question answering, and artificial intelligence.

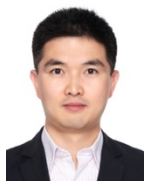

**Yongshuai Hou**, PhD, is an algorithm engineer at Peng Cheng Laboratory, Shenzhen, China. His research interests cover natural language processing, question answering system and medical informatics.

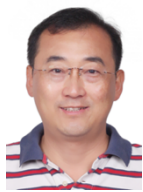

**Ge Li** (Member, IEEE) is currently a Professor with the School of Electronic and Computer Engineering, Shenzhen Graduate School, Peking University, China. His research interests include image/video process and analysis, machine learning, digital communications, and signal processing.

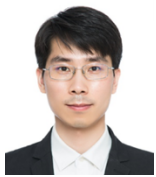

**Hui Wang**, M.S. degree of Harbin Institute of Technology，now is a leader of the algorithm team of Gennlife (Beijing) Technology co., Ltd. His research interests cover artificial intelligence, machine learning and medical informatics

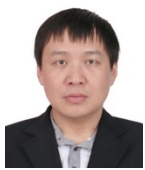

**Hui Xu**, PhD of Peking University, now is chief technology officer of Gennlife (Beijing) Technology co., Ltd. His research interests cover artificial intelligence, machine learning and medical informatics

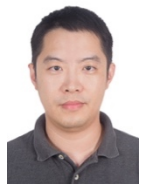

**Yang Xiang** (Member, IEEE), PhD, is an assistant professor at Peng Cheng Laboratory, Shenzhen, China. His expertise is in natural language processing, medical informatics and clinical artificial intelligence.

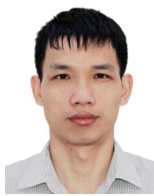

**Buzhou Tang** received the Ph.D. degree from the Harbin Institute of Technology, China, in 2011. He is currently an associate Professor with Harbin Institute of Technology, Shenzhen, China, and Peng Cheng Laboratory, Shenzhen, China. His research interests cover artificial intelligence, machine learning, medical informatics, natural language processing and signal processing.